\documentclass[10pt, a4paper]{article}

\usepackage{lrec-coling2024} %
\usepackage{booktabs}
\usepackage{makecell}
\usepackage{stackengine}
\usepackage{xcolor}
\usepackage{microtype}

\title{Dataset of Quotation Attribution in German News Articles}

\name{Fynn Petersen-Frey, Chris Biemann} 

\address{House of Computing and Data Science \& Language Technology Group \\
          Universität Hamburg \\
         \texttt{\{fynn.petersen-frey, chris.biemann\}@uni-hamburg.de}\\}

\abstract{
Extracting who says what to whom is a crucial part in analyzing human communication in today's abundance of data such as online news articles.
Yet, the lack of annotated data for this task in German news articles severely limits the quality and usability of possible systems.
To remedy this, we present a new, freely available, creative-commons-licensed dataset for quotation attribution in German news articles based on WIKINEWS.
The dataset provides curated, high-quality annotations across 1000 documents (250,000 tokens) in a fine-grained annotation schema enabling various downstream uses for the dataset.
The annotations not only specify who said what but also how, in which context, to whom and define the type of quotation.
We specify our annotation schema, describe the creation of the dataset and provide a quantitative analysis.
Further, we describe suitable evaluation metrics, apply two existing systems for quotation attribution, discuss their results to evaluate the utility of our dataset and outline use cases of our dataset in downstream tasks.
 \\ \newline  }

\newcommand{\cue}[1]{\textit{Cue}}
\newcommand{\adr}[1]{\textit{Addressee}}
\newcommand{\msg}[1]{\textit{Quote}}
\newcommand{\msgs}[1]{\textit{Quotes}}
\newcommand{\src}[1]{\textit{Speaker}}
\newcommand{\frm}[1]{\textit{Frame}}

\newcommand{\dir}{\textit{Direct}}
\newcommand{\indi}{\textit{Indirect}}
\newcommand{\rep}{\textit{Reported}}
\newcommand{\frin}{\textit{Free Indirect}}
\newcommand{\infr}{\textit{Indirect/Free Indirect}}

\newcommand{\speech}{\textit{Speech}}
\newcommand{\writing}{\textit{Writing}}
\newcommand{\thought}{\textit{Thought}}
\newcommand{\sw}{\textit{Speech/Writing}}
\newcommand{\spt}{\textit{Speech/Thought}}
\newcommand{\wt}{\textit{Writing/Thought}}

\newtheorem{exmp}{Example}[section]

\newlength\lunderset
\newlength\rulethick
\def\stackalignment{l}
\newcommand\nunderline[3][1]{\setbox0=\hbox{#2}%
  \stackunder[#1\lunderset-\rulethick]{\strut#2}{\color{#3}\rule{\wd0}{\rulethick}}}
\newcommand\nundertext[2][1]{\def\useanchorwidth{T}\def\stackalignment{c}\smash{%
  \stackunder[#1\lunderset-\rulethick+.6\lunderset]{}{%
  \scriptsize\strut#2}}}
\newcommand{\cueline}[2][0]{\nunderline[#1]{#2}{red}}
\newcommand{\cuetext}[1][0]{\nundertext[#1]{\cue}}
\newcommand{\addr}[2][0]{\nunderline[#1]{#2}{violet}}
\newcommand{\addrtext}[1][0]{\nundertext[#1]{\adr}}
\newcommand{\speaker}[2][0]{\nunderline[#1]{#2}{blue}}
\newcommand{\speakertext}[1][0]{\nundertext[#1]{\src}}
\renewcommand{\frame}[2][1]{\nunderline[#1]{#2}{cyan}}
\newcommand{\frametext}[1][1]{\nundertext[#1]{\frm}}
\newcommand{\direct}[2][0]{\nunderline[#1]{#2}{green}}
\newcommand{\directtext}[1][0]{\nundertext[#1]{\dir}}
\newcommand{\indirect}[2][0]{\nunderline[#1]{#2}{orange}}
\newcommand{\indirecttext}[1][0]{\nundertext[#1]{\indi}}
\newcommand{\reported}[2][0]{\nunderline[#1]{#2}{purple}}
\newcommand{\reportedtext}[1][0]{\nundertext[#1]{\rep}}
\newcommand{\freein}[2][0]{\nunderline[#1]{#2}{brown}}
\newcommand{\freeintext}[1][0]{\nundertext[#1]{\frin}}
\newcommand{\indifree}[2][0]{\nunderline[#1]{#2}{pink}}
\newcommand{\indifreetext}[1][0]{\nundertext[#1]{\infr}}

\begin{document}

\maketitleabstract

\section{Introduction}

Ever-increasing amounts of data including discourses in natural language are produced in today's digital era.
When scientists or journalists want to analyze this data such as online news articles, they are facing the issue that it is infeasible to manually work through the enormous amounts of data.
Extracting who says what to whom is a crucial part in analyzing human communication, how the discourse changes over time or what quotations are reproduced by which media etc.
Although the field of natural language processing has made huge leaps forward with the introduction of transformers, there is no suitable, annotated data to train a transformer-based system to extract who said what to whom for modern German news articles. %

In this paper, we present a creative-commons-licensed dataset for quotation attribution in German news articles.\footnote{Available at \url{https://github.com/uhh-lt/german-news-quotation-attribution-2024}}
The dataset consists of 1000 manually annotated articles from the German WIKINEWS website\footnote{URL: \url{https://de.wikinews.org}}.
In total, these annotated articles contain almost 250,000 tokens.
We manually annotated and curated \msgs{} in different forms of speech such as \dir, \indi, \frin,  \infr, \rep{} together with the corresponding \frm{}, \src{}, \cue{} and \adr{}.

An overview of the span annotation classes can be found in Table \ref{tab:classes}. This includes short descriptions, number of occurrences in the data and their average length.
Table \ref{tab:types} gives an overview of quote types including short descriptions, the number of occurrences and average length.
In addition, we provide a number of annotated sentences in Examples \ref{ex:dir}--\ref{ex:infr} to get a quick intuition on the dataset and its annotations.
These examples are modeled after cases from the curated dataset.
We shortened or changed the content as needed to be presentable in this text while keeping the structure and grammatical phenomenon as it was.

\begin{table*}
  \begin{tabularx}{\textwidth}{lXrr}
    \toprule
    annotation & short description                                                                 & count & avg. len. \\
    \midrule
    \msg{}     & the quotation uttered by the \src{}, fine-grained labels in Table \ref{tab:types} & 4182  & 16.69     \\
    \src{}     & entity in the text that utters the quotation                                      & 3908  & 3.53      \\
    \cue{}     & words that are part of a \frm{} and signal a \msg{} construction                  & 2929  & 1.57      \\
    \frm{}     & part of a sentence including \cue{} \& \src{}, but not the quotation              & 3038  & 8.95      \\
    \adr{}     & entity in the text that the quotation is directed at                              & 337   & 2.72      \\
    \bottomrule
  \end{tabularx}
  \caption{Overview of quotation attribution spans}
  \label{tab:classes}
\end{table*}

\begin{table*}
  \begin{tabularx}{\textwidth}{lXrr}
    \toprule
    type    & short description                                                                                                    & count & avg. len. \\
    \midrule
    \dir{}  & actual words of an utterance, usually in quotation marks                                                             & 873   & 17.54     \\
    \indi{} & content-wise equivalent utterance using different words, usually part of a sentence together with a \frm{}           & 2250  & 14.71     \\
    \rep{}  & report of a speech action, possibly far from the original quote, usually a full sentence, no \frm{}                  & 454   & 18.01     \\
    \frin{} & mix of article author \& actual speaker, typically construct with "sollen" (shall) or "müssen" (must), full sentence & 171   & 20.42     \\
    \makecell[tl]{\textit{Indirect/}                                                                                                                   \\\textit{Free Indirect}} & content-wise equivalent utterance written in conjunctive mood, full sentence & 434 & 22.33 \\
    \bottomrule
  \end{tabularx}
  \caption{Overview of the quotation types}
  \label{tab:types}
\end{table*}

\begin{exmp}[\dir{}]
  \item  \frame{\cueline{Zit\cuetext{}at} von\frametext{} \speaker{Mer\speakertext{}kel}:} \direct{„Wir scha\directtext{}ffen das.“}\vspace{0.75em}
  \item \em \frame{\cueline{Quo\cuetext{}te} from\frametext{} \speaker{Mer\speakertext{}kel}:} \direct{„We can\directtext{} do this.“}\vspace{0.75em}
  \label{ex:dir}
\end{exmp}

\begin{exmp}[\indi{}]
  \item  \frame{\addr{Der Nachricht\addrtext{}enagentur} \cueline{s\frametext{}a\cuetext{}gte} \speaker{e\speakertext{}r},} \indirect{dass m\indirecttext{}an eine} \indirect{Lösung finden werde}.\vspace{0.75em}
  \item \em \frame{\speaker{He\speakertext{}} \cueline{told\cuetext{}} \addr{the \frametext{}news\addrtext{} agency}} \indirect{that a solut\indirecttext{}ion would be} \indirect{found}.\vspace{0.75em}
  \label{ex:indi}
\end{exmp}

\begin{exmp}[\rep{}]
  \item \reported[1]{\speaker{Die\speakertext{} Firma} forderte eine schnellere\reportedtext[1]{} Entscheidung.}\vspace{0.75em}
  \item \em \reported[1]{\speaker{The\speakertext{} company} demanded a quicker\reportedtext[1]{} decision.}\vspace{0.75em}
  \label{ex:rep}
\end{exmp}

\begin{exmp}[\frin{}]
  \item \speaker{Ein Sp\speakertext{}recher} stellte gestern die neuen Ziele vor.\vspace{0.5em} \freein{Es soll mehr Geld in\freeintext{} die Bildung fließen.}\vspace{0.75em}
  \item \em \speaker{A spok\speakertext{}esman} presented the new goals yesterday.\vspace{0.5em} \freein{More money is to \freeintext{}flow into education.}\vspace{0.75em}
  \label{ex:frin}
\end{exmp}

\begin{exmp}[\infr{}]
  \item \speaker{Ein Pa\speakertext{}ssant} schilderte die Situation. \indifree{Die Polizei}\vspace{0.75em} \indifree{habe den Bereich  \indifreetext{}großräumig abgeriegelt.}\vspace{0.75em}
  \item \em \speaker{A pa\speakertext{}sserby} described the situation. \indifree{The police}\vspace{0.75em} \indifree{had cordoned off the\indifreetext{} area over a wide area.}\vspace{0.75em}
  \label{ex:infr}
\end{exmp}

In the following, we review related work on quotation detection and attribution before describing our annotation schema.
Then, we describe the creation of our dataset and perform experiments including a quantitative analysis as well as an application of two existing systems for quotation attribution.
Before concluding, we describe use cases for our dataset.

\section{Related Work}
\label{sec:relwork}

The task of quotation detection and attribution to a speaker has been tackled by numerous approaches, usually with the goal to extract information from the data such as news articles \citep{krestel-etal-2008-minding,pareti-etal-2013-automatically,almeida-etal-2014-joint,scheible-etal-2016-model}.
Earlier approaches were purely rule-based and only dealt with quotation detection.
More recent works used data-driven methods (often based on the PARC dataset by \citet{pareti-2012-database}) to detect quotations and their respective speakers.
However, there is still a strong focus on resources for direct quotations such as the software by \citet{pouliquen:etal:2007} or the datasets by \citet{okeefe-etal-2012-sequence} and \citet{zhang-liu-2022-directquote} while only few resources  provide indirect quotations as they are not as easily extracted automatically.

A similar task is speaker attribution in literary works.
As the literary domain differs from the news domain e.g. in the author perspective, in the type of quotations and the focus on characters as implicit or explicit speakers, the field of computational literary studies has seen numerous studies dealing with speaker attribution in literary works \citep{elson-etal-2010-extracting,he-etal-2013-identification,muzny-etal-2017-two}.

While many works have addressed quotation detection and attribution in English, less work and resources have been created for other languages.

For historical German texts,
\citet{Brunner2015,krug:etal:2018}, \citet{brunner-etal-2019-deep} and \citet{BrunnerEngelbergJannidisetal.2020} have created a number of resources.
The DROC corpus \citep{krug:etal:2018} consists of 90 fragments of German novels and includes about 2000 annotated direct quotes and annotations for speakers and addresses.
The Redewiedergabe corpus \citep{BrunnerEngelbergJannidisetal.2020} extends this work by creating a historical corpus (mostly literary domain, but also some news articles) with fine-grained annotations for speech, thought and writing.
\citet{BogelG2015} created a system to extract statements from German news articles.
For Finnish, \citet{janicki-etal-2023-detection} recently created an annotated dataset of news articles with quotations and their speakers.

Our research is focused on who says what to whom according to German news media.
Since no suitable dataset exists for this purpose, we created a new manually annotated and curated dataset for quotation attribution in German news articles.

\section{Annotation Schema}
\label{sec:annotation-schema}

The annotation schema is inspired primarily by the Redewiedergabe project \citep{BrunnerEngelbergJannidisetal.2020} and also by the work of \citet{BogelG2015}.
We annotate five different (possibly discontinuous) spans:
Beside the actual quotation annotated as \msg{}, spans of \src{}, \cue{}, \frm{} and \adr{} (not part of the Redewiedergabe project) are annotated as optional roles for each \msg{}.
For a \msg{} two additional dimensions are coded: Five different types of speech (\dir{}, \indi{}, \rep{}, \frin{}, \infr{}) as well as six media (\speech{}, \thought{}, \writing{}, and not part of the Redewiedergabe project \spt{}, \sw{}, \wt{}).
This produces 30 different combinations of quotations.
Examples \ref{ex:dir}--\ref{ex:infr} provide short sentences showing the annotations.
While we reused some class names from the Redewiedergabe project and tried to align our classes, we modified definitions from the Redewiedergabe annotation guidelines or created new definitions suited for news articles and nested quotations.
In the next sections, we define the roles, types and media.

\subsection{Roles}

A quotation in itself is of little value when it is not known who said it or in what context the statement was made.
Thus, we provide roles as additional spans that are linked to one or more quotations.
The \src{} identifies who said something, the \cue{} describes how it was uttered (possibly negating/adversary!) and the \frm{} provides context so that a quotation is not free-floating.

\paragraph{\src{}}
\label{sec:src}
The speaker is the linguistic phrase in the text that utters one or more \msgs{}.
This is typically a personal pronoun, named entity or a noun phrase subject in a sentence.
Explanatory relative clauses as well as content clauses (clauses with \textit{dass} (that) or \textit{ob} (whether) making an attribution to a subject) are not annotated.
However, noun phrase modifiers (explanatory attributions of subjects) are annotated (e.g. \textit{\speaker{der} \speaker{zufälligerweise anwesende Doktor}} (the doctor who happens to be present) is the full \src{}.
Usually, the speaker of \dir{} and \indi{} speech is located within the associated frame.
For \rep{}, the speaker can be found within the quotation span.
The \src{} of \frin{} and \infr{} is outside the respective quotation span.

\paragraph{\cue{}}
\label{sec:cue}
A \cue{} consists of signal words in a frame that announce a \dir{} or \indi{} speech. These are usually verbs. However, they can also be specified expressions (e.g.: \textit{\cueline{laut}, \cueline{nach}, \cueline{so}, \cueline{zufolge}} (according to).
The \cue{} span can also be split within a frame (typically for German verb pre- or suffixes).
Besides the reflexive pronoun, a verb can also include other parts of speech such as prepositions, adverbs, nouns and adjectives which distinguish the verb from similar verbs; e.g.: \textit{\cueline{von} \dots \cueline{die Rede  sein}} (talk of \dots), \textit{\cueline{für wahrscheinlich halten}} (consider likely).

\paragraph{\adr{}}
The \adr{} is the linguistic phrase in the text that a quotation is directed at.
It is typically found within a \frm{} or within the quotation span in case of \rep{}.

\paragraph{\frm{}}
The part of a sentence that is outside the quotation and contains \cue{}, \src{} and possibly \adr{}. The frame provides context for the quotation. It can be at the beginning, in the middle or at the end of a sentence. It is also possible to split the \frm{} within a sentence if it is interrupted by the quotation.
The \frm{} in \indi{} and \dir{} speech is usually a clause, annotated with its comma or colon, which separates the quotation from the \frm{}.

\subsection{Quotation Types}

Quotations come in various forms and shapes.
These differ in their level of truthfulness of the reproduction to the original utterance.
To account for this, we marked a span of text not only as a \msg{} but also labeled it according to the five classes described in the following paragraphs.
This additional information per quotation allows to use the dataset for downstream tasks that are only interested in a specific type of quotation.
When further processing extracted quotations, systems can consider the reproduction truthfulness in their methods and differentiate e.g. between a \dir{} quotation and only vaguely related \rep{} speech.

\paragraph{\dir{}}
The \dir{} label is used for verbatim reproductions of quotations that usually occur enclosed in quotation marks.
Typically, it either directly follows an introductory \frm{} or the associated \frm{} immediately follows the quotation.
\frm{} and \dir{} speech can be separated either by colons or commas that belong to the \frm{}.
In other cases, fragments of \dir{} quotations are integrated into a sentence and cannot stand on their own.
Then, the \dir{} speech is nested in a longer quotation of any of the remaining four quotation types.

\paragraph{\indi{}}
A quotation is labeled as \indi{} whenever the author of a text indicates in an associated \frm{} that the utterances of another person are reproduced as a paraphrase, not verbatim.
It is the only type of \msg{} that always requires a \frm{}.
The usual type of \indi{} speech is a partial sentence that, together with a \cue{} in the \frm{}, forms the complete sentence.
In a special case, the \cue{} is a single word reference (e.g. \textit{wonach, demnach, danach} whereupon, thus) to a \src{} in the previous sentence.

\paragraph{\rep{}}
A \rep{} quotation is a summary report of a statement made by another person that is reproduced in a free manner possibly far from the original statement.
Because a reporting style is common in news texts, \rep{} speech is annotated only in cases where, first, there is a clearly identifiable \src{} within the quotation span, and second, the quotation contains information uttered by the \src{} -- not only a description of an action.

\paragraph{\frin{}}
A quotation is labeled \frin{} when statements, writings or thoughts of a person are reproduced who is not the article author, but the quotation is nevertheless written from the author's perspective.
Mostly formulations with \textit{sollen} (shall) and \textit{müssen} (must) that reflect foreign thoughts, statements or writings are considered. A \frin{} quotation is usually a complete sentence that is not enclosed by quotation marks, nor does it have a \frm{} in the same sentence.
However, it has a \src{} that is outside the quotation span and thus also outside the sentence.

\paragraph{\infr{}}
This class is used for those forms of speech reproduction which, without an introductory \frm{}, reproduce statements of another entity in a sentence in the subjunctive mood, but not verbatim.
An \infr{} quotation occurs only when any of the other four types of quotation occur in the preceding or succeeding sentence that also provide a \src{}.

\subsection{\msg{} Media}

The media of speech reproduction indicate whether a quotation is a \speech{}, \thought{} or \writing{} action.
These three media are only annotated if it can be clearly determined which particular medium was used in a quotation.
If the media cannot be clearly determined or two different media apply at the same time, the mixed forms are chosen.

\paragraph{\speech{}}
A \msg{} is labeled \src{} when the reproduced utterance was originally oral. This medium of speech reproduction is often recognizable in newspaper texts by \cue{} verbs that are unambiguous for a spoken reproduction; e.g.: \textit{sagen} (say), \textit{sprechen} (speak). But also the \src{} can give clarity about a spoken utterance; e.g. \textit{der Sprecher} (the speaker).

\paragraph{\thought{}}
The \thought{} class marks a reproduced cognitive process where the statement originally occurred mentally.
Consequently, \thought{} is rarely found in news articles as the author cannot reproduce the thoughts of another person.

\paragraph{\writing{}}
This class labels a reproduced writing process or a written form of language.
Similar to \speech{}, the \cue{} or \src{} can be indicators.

\paragraph{\spt{}}
This mixed medium is used to label a person's oral statement in which they have expressed their thoughts.

\paragraph{\sw{}}
This mixed medium marks a \msg{} a) when it's uncertain whether the original quote comes from a written or oral source or b) if a quotation is made as a combination of texts as well as oral statements.

\paragraph{\wt{}}
This mixed medium is chosen when a person's writing is cited in which he or she has reproduced his or her thoughts. Our dataset of news articles has no instances of this class.

\section{Dataset Creation}

After describing the annotation schema,  we provide details on the source data, its pre-processing, the annotation process, the inter-annotator agreement and handling of disagreement between annotators in the following sections.

\subsection{Source Data}
The data originates from news articles published on the German WIKINEWS website.
We used the XML dump\footnote{URL: \url{https://dumps.wikimedia.org/dewikinews/}} available through the Wikimedia foundation.
Our dataset is based on the dump from April 2022 that consists of 13,001 published articles.
From these published articles, we randomly sampled 1000 articles for annotation to stay close to the original distribution while reducing the data size to an amount manageable in our project timeframe.
These articles range from December 2004 to March 2022.

\subsection{Data Pre-Processing}
As articles stored in MediaWiki markup contain custom macros for the German WIKINEWS, we wrote a program to obtain plain text.
The conversion is a recursive procedure to support the nested macros present in the markup.
Using this approach, we stripped all markup like formatting (e.g. bold, italic), semantic information (e.g. links to entities on Wikipedia) and non-textual content (e.g. pictures, tables) from the documents.
Further, we removed any text not belonging to the main text body such as publication metadata, comments, links to related articles or sources.
The resulting plain text was tokenized and split into sentences using spaCy \citep{honnibal2020spacy}.

\subsection{Annotation Process}
\label{sec:annotation-process}

The annotation was carried out by three annotators with a background in German studies or Linguistics.
The annotators were selected after performing a trial annotation on a handful of articles.
The annotation team received extensive training during a preliminary annotation before the actual annotation begun.
Further, we held weekly meetings during the main annotation to discuss open questions and uncertain cases, thereby providing ongoing training to all annotators.
The annotation quality of our annotators did not differ in a noticeable way after training.
Neither in the discussions nor in the curation did it became evident that the annotations of one annotator were preferred over annotations of another annotator.

In an initial preliminary annotation, we tested the suitability of the annotation schema in the news domain.
We iteratively tested which attributes of the schema are necessary and which additional options we needed.
Finally, we settled on the medium and type attribute for a \msg{} and \frm{}, \cue{}, \src{} and \adr{} as the other annotation components (roles).

For the annotation, we used the annotation software INCEpTION \citep{klie-etal-2018-inception}.
The different components are modeled as span annotations with relations between them to indicate e.g. which \src{} belongs to which \msg{}.
We divided our sampled documents into six parts to a) annotate and curate in parallel, b) allow to adapt the annotation schema early in the process if needed and c) track the inter-annotator agreement over time.
We decided against automatic highlighting of candidate annotations etc. to not introduce any automatic processing bias.
Thus, the instances were always manually identified by searching for the suitable grammatical structures.

\subsection{Inter-Annotator Agreement}
\begin{table}
  \centering
  \begin{tabular}{lrrrr}
    \toprule
    {part} & {type} & {medium} & {roles} \\
    \midrule
    Part 1 & 0.56   & 0.37     & 0.61    \\
    Part 2 & 0.76   & 0.51     & 0.75    \\
    Part 3 & 0.77   & 0.40     & 0.76    \\
    Part 4 & 0.77   & 0.68     & 0.76    \\
    Part 5 & 0.86   & 0.51     & 0.83    \\
    Part 6 & 0.78   & 0.61     & 0.78    \\
    \bottomrule
  \end{tabular}
  \caption{Krippendorff's Alpha agreement between the annotators on the six parts}
  \label{tab:iaa}
\end{table}
We use Krippendorff's Alpha to compute the agreement between two annotators per part.
The measure includes both the quality of the span annotation offsets (overlap) and their labels, but does not include the relations between the span annotations.
However, the relations were typically made identical given the same annotation spans and their labels.
Moreover, for different annotation spans, there is no sensible way to compute an inter-annotator agreement on the relations.

Table \ref{tab:iaa} shows the inter-annotator agreement values for the six parts into which we divided the 1000 documents.
The inter-annotator agreement values increased strongly after the first part, slightly increasing with additional experience and training over the course of the remaining parts.
As such, the first part required significant curation effort and discussion that ultimately led to improved skills of our annotators.
The inter-annotator agreement values for the medium fluctuate and show the lowest numbers in general because annotating the correct medium proved to be difficult depending on the context.
The documents in part 3 (with the drop to 0.4) had many quotations where the medium was challenging for the annotators to select.

Neither the Redewiedergabe project nor \citet{BogelG2015} report inter-annotator agreement scores to compare to.
As our annotation is a lot more complex than most span-based annotations (e.g. named-entity recognition) it is to be expected that our scores a lower.
With levels around 0.76 for type and roles, the scores are only slightly lower than the typical scores achieved in simpler span annotations tasks.

\subsection{Disagreements between Annotators}

During the annotation phase we held weekly meetings to discuss general questions how would we best annotate a specific phenomenon within our annotation schema.
After two annotators had finished annotating the documents, we employed curation by a third person to resolve differences in the annotations.
In situations where the curator was not certain who (or if any) of the two annotators  had correctly annotated the sentences in question, we discussed the issue in detail to resolve the disagreement, thereby potentially defining our annotation guidelines more precisely.

One of the most frequent reasons of disagreement during the early phases of the annotation was the difficulty of choosing the correct medium, usually the choice was between \writing{} or \speech{}.
After many discussions, we concluded that it is sometimes impossible to decide from the text alone whether an utterance was produced in spoken or written form.
As such, we modified our annotation schema by adding three new labels to medium.
While this increased the annotation consistency considerably, it did not completely resolve the issue as the inter-annotator agreement shows.

\subsection{Final Dataset}
We exported and converted the curated articles into a JSON representation.
During the conversion, we applied automatic checks for potential annotation errors and manually resolved true errors in the curated documents.
The relations between the annotation spans allow us to build tuples where each tuple consists of one quotation with type and medium as well as all linked roles.
Each text span is provided with character, token and sentence offsets enabling easy usage in various NLP frameworks.

\section{Experiments}

In this section, we present the experiments we performed on the dataset.
First, we conduct a quantitative analysis of the annotations.
Second, we evaluate two systems on the dataset after explaining the systems and defining evaluation metrics.

\subsection{Dataset Analysis}
\label{sec:analysis}

In this section, we provide a quantitative view of the annotations in our dataset.
The total count and average length of each \msg{} type is shown in Table \ref{tab:types}.
Table \ref{tab:classes} provides the equivalent data for the role annotations.
While most \msgs{} have a \src{}, only 70\% have a \cue{} or \frm{}.

\begin{figure}
  \centering
  \includegraphics[width=\linewidth]{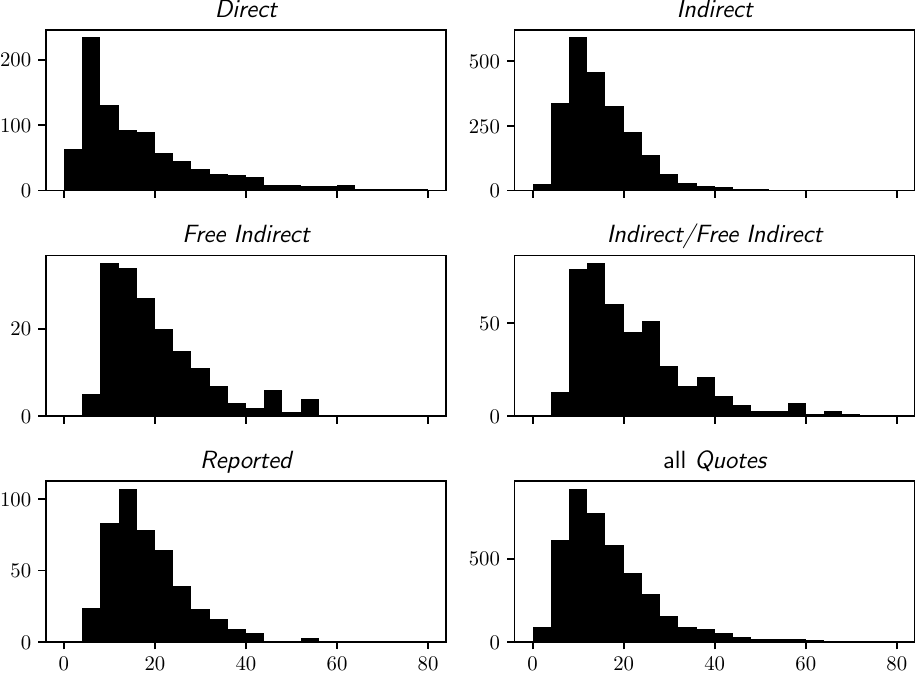}
  \caption{\msg{} token length histograms}
  \label{fig:quoteHist}
\end{figure}
Figure \ref{fig:quoteHist} shows histograms of the token lengths for the different types of quotations.
Overall, quotations lengths approach a heavily skewed normal distribution with some very short spans and a long tail for rare, long spans.
Most quotations consist of 5 to 20 tokens.
\dir{} quotations contain both the shortest and the longest quotations in the dataset.
\indi{} quotations are the shortest on average as they are usually fragments in a single sentence.
\rep{}, \frin{} and \infr{} (in increasing order of average length) have similar distributions leaning towards longer spans since they normally consist of at least one but sometimes a few sentences.

\begin{figure}
  \centering
  \includegraphics[width=\linewidth]{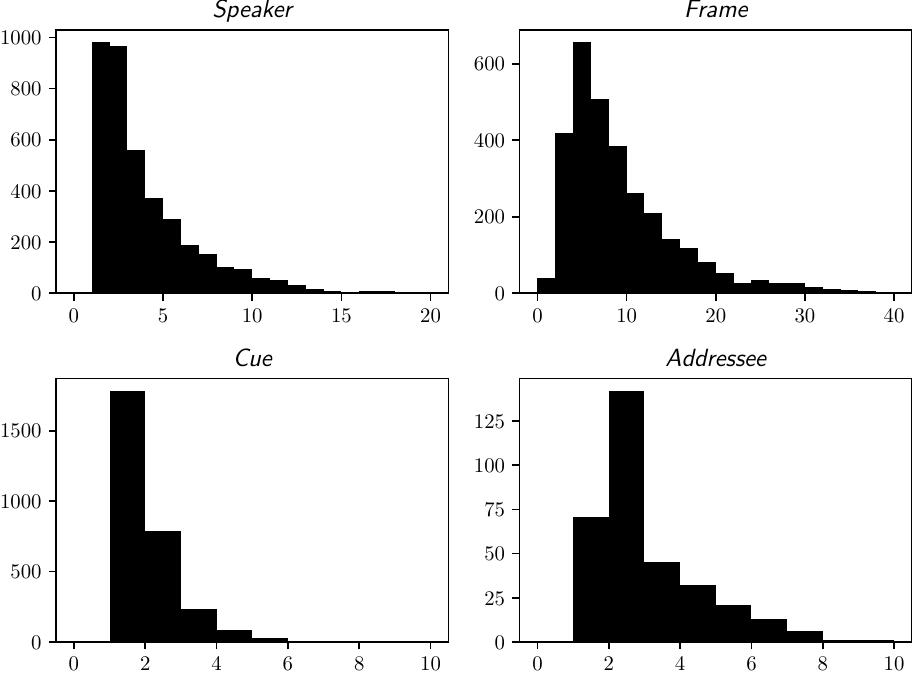}
  \caption{Role token length histograms}
  \label{fig:roleHist}
\end{figure}
Figure \ref{fig:roleHist} shows the equivalent histograms for the roles.
\src{} follows a Poisson distribution where most speaker spans are shorter than 5 tokens.
Yet, some \src{} spans include descriptive phrases leading to more than 15 tokens (see Section \ref{sec:src} for details).
The \frm{} annotations follow a skewed normal distribution like the \msgs{}.
Since they can be a full sentence in length, they are the longest of the role spans with up to 40 tokens.
\cue{} spans are the shortest annotations; typically a single token.
However, around 33\% of the \cue{}\textit{s} are multi-token expressions.
The \adr{} is a very rare annotation with lengths between one and eight tokens.

\begin{figure}
  \centering
  \includegraphics[width=\linewidth]{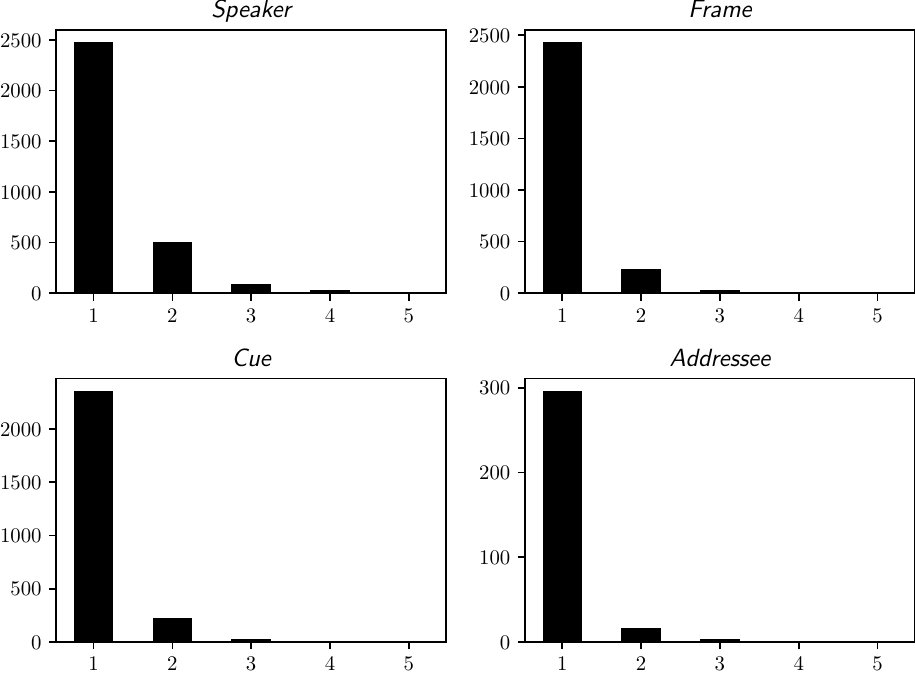}
  \caption{Number of \msgs{} per role span}
  \label{fig:roleUses}
\end{figure}
Figure \ref{fig:roleUses} shows the number of quotations each role annotation is attributed to.
In the overwhelming majority of cases each role span is only used for a single quotation.
However, some spans are attributed to two \msg{} spans.
This is especially true for the \src{} where up to five quotations are attributed to a single \src{} annotation.

We further analyzed the amount of nested \msgs{}, the number of sentences in a \msg{} and the distance between a role and its corresponding \msg{}.
10\% of all \msgs{} are nested inside another \msg{} or \frm{}.
The majority of these cases are instances of \dir{} speech fragments.
While most \msgs{} span only one sentence, about 10\% span two or more sentences (up to 11 sentences).
11\% of all role spans and 21\% of all \src{} spans are one or more sentences apart from the corresponding \msg{} (up to 7 sentences).

From these quantitative observations it becomes apparent that a system requires the following traits to be able to perform well:
1) Find \msgs{} and match the \src{} without any \cue{}.
2) Support for multi-word \cue{} spans.
3) Support \msg{} spans over multiple sentences.
4) Support finding roles for a \msg{} in other sentences.
5) Find \indi{} and \dir{} quotations, also \rep{} and \infr{} as they account for 10\% each, while \frin{} makes up only 4\% of all \msgs{}.
6) Support role spans to be used for multiple \msgs{}.
7) Support nested \msg{} spans.

\subsection{Baseline Systems}

In order to evaluate the utility of our dataset, we apply two baseline systems that can extract quotations with attributed roles from news articles.
The first system is a purely rule-based system that does not need any training data.
The second system uses a data-driven machine learning approach.

An apparently similar problem to the annotations in our dataset is semantic-role-labeling (SRL).
However, we did not evaluate SRL systems because the typical SRL datasets and thus available systems are limited to work on a single sentence as a unit.
This is not suitable for our dataset that requires a document-wide context (or at least multi-sentence context) as quotes span multiple sentences and roles appear in yet other sentences.

\paragraph{Rule-based system}
We developed a rule-based system (RBS) building on top of spaCy \citep{honnibal2020spacy} to extract direct and indirect quotations with the speaker from text.
The system follows ideas of an older system presented by \citet{BogelG2015}.
It uses rules and word lists on top of neural components for dependency parsing and named-entity recognition.
\dir{} speech is identified by regular expressions looking for quotation marks.
The \src{} of the quotation (i.e. the speaker) is searched in the proximity, preferring candidates in the same sentence but outside the quotation span.
\indi{} speech is identified through the grammatical structure of a sentence (using dependency parsing) and the main or auxiliary verb being a cue word that is looked up in a word list.
The word list contains utterance verbs (\textit{verba dicendi}) that can be used to indicate (in)direct speech.
In addition, the system finds sentences in subjunctive mood that occur directly before or after a sentence containing another quotation.
These sentences are typically marked as \infr{} in the dataset.
Lastly, the system combines \dir{} and \indi{} speech, enriching the information of identical quotations.
The system does not handle the \adr{} span.
Since it is a rare class, we simply ignore it and do to not predict any \adr{}.
However, \frm{} is a frequent role that the system predicts by marking all tokens of a sentence as the \frm{} that do not belong to the \dir{} or \indi{} quotation.

\paragraph{Citron}
The system was created by the BBC \citep{citron} to extract quotations with their \src{} from English news articles.
It consists of several components that are built on top of spaCy \citep{honnibal2020spacy} and are trained individually: \cue{} classifier, \src{} classifier, \src{} resolver, \msg{} classifier, \msg{} resolver.
The resolvers link the classified spans to a \cue{}.
The system can only find quotations that have a \cue{} -- with the additional constraint that a \cue{} is single token verb.
We modified Citron to work with German texts, accept any single token as a \cue{} and trained it using the subset of all quotations that have a \cue{} in our dataset (70\%).
As with RBS,
we ignore any \adr{} and predict the \frm{} to span all tokens in a sentence not belonging to the \msg{}.
To predict the quotation type, we use \dir{} for spans with quotations marks and \indi{} otherwise.

\subsection{Evaluation Metrics}
We use the usual precision, recall and F1-metrics on token overlap of possibly discontinuous spans (thereby creating ordered sets of tokens).
For most \msg{} types, all roles are optional.
Thus, predicted spans of roles can only be matched to the reference roles if they belong to a correctly matched \msg{}.
A span representing a role can be related to multiple \msg{} spans, i.e. the same \src{} can utter multiple \msgs{}.
Roles or \msg{} spans can be nested within another \msg{} or \frm{}.
To perform an evaluation, \msgs{} from system and reference are assigned via linear sum assignment of the \msg{} span's token overlap using type and medium as tie-breakers.
Each \msg{} can only be matched to at most one other \msg{}.
The tie-breakers are needed to correctly assign \msgs{} in rare cases as they can have the same offsets, yet are of a different type or medium.
If a system predicts a \msg{} that has no matching \msg{} in the reference annotations, this increases the false positives for \msg{} and each role the system predicted as belonging to the unmatched \msg{}.
Vice versa, if a \msg{} from the reference annotation has no match in the system prediction, the false negatives are increased.
A correctly matched \msg{} yields true positives for all correct roles according to the fraction of overlap and false negatives resp. false positives for tokens that were not identified resp. wrongly predicted by the system.

\subsection{Results}
\label{sec:results}

\begin{table*}[tbhp]
  \begin{center}
    \begingroup
    \setlength{\tabcolsep}{4.8pt}
    \begin{tabular}{llcccccccccccc}
      \toprule
             &            & \multicolumn{3}{c}{quotation} & \multicolumn{3}{c}{roles} & \multicolumn{3}{c}{joint} & \multicolumn{3}{c}{type}                                                           \\
      \cmidrule(lr){3-5}
      \cmidrule(lr){6-8}
      \cmidrule(lr){9-11}
      \cmidrule(lr){12-14}
      system & data       & prec.                         & rec.                      & F1                        & prec.                    & rec. & F1   & prec. & rec. & F1   & prec. & rec. & F1   \\
      \midrule
      RBS    & dev        & 75.1                          & 36.1                      & 48.8                      & 55.0                     & 25.5 & 34.9 & 60.7  & 28.7 & 38.9 & 57.8  & 29.6 & 39.1 \\
      RBS    & test       & 70.8                          & 36.2                      & 47.9                      & 55.6                     & 26.1 & 35.5 & 59.9  & 29.0 & 39.1 & 63.5  & 33.6 & 43.9 \\
      Citron & dev        & 91.5                          & 27.6                      & 42.4                      & 79.3                     & 31.5 & 45.1 & 82.4  & 30.3 & 44.3 & 87.0  & 26.6 & 40.8 \\
      Citron & test       & 88.2                          & 30.1                      & 44.9                      & 77.9                     & 34.2 & 47.6 & 80.5  & 33.0 & 46.8 & 86.5  & 29.6 & 44.1 \\
      \midrule
      Citron & dev cue    & 88.4                          & 39.9                      & 55.0                      & 80.1                     & 37.1 & 50.7 & 82.2  & 37.8 & 51.8 & 91.5  & 41.1 & 56.7 \\
      Citron & dev 1 cue  & 79.0                          & 44.6                      & 57.0                      & 73.5                     & 47.1 & 57.4 & 74.9  & 46.5 & 57.3 & 82.1  & 48.6 & 61.1 \\
      Citron & test cue   & 85.9                          & 41.6                      & 56.1                      & 80.6                     & 40.3 & 53.7 & 81.9  & 40.6 & 54.3 & 90.1  & 42.9 & 58.2 \\
      Citron & test 1 cue & 80.0                          & 47.6                      & 59.7                      & 74.7                     & 48.3 & 58.8 & 76.0  & 48.3 & 59.0 & 85.4  & 50.7 & 63.6 \\
      \bottomrule
    \end{tabular}
    \endgroup
  \end{center}
  \caption{Evaluation results}
  \label{tab:results}
\end{table*}

To evaluate the two baseline systems, we divided our dataset into three parts:
A training set of 700 documents, 150 documents for the development set (653 quotations, 1567 roles) and 148 documents in the test set (652 quotations, 1605 roles).
Table \ref{tab:results} (upper half) shows the results for the two baseline systems on the development/test set.
We do not report scores on the medium because neither system is capable to predict it.
The rule-based system is not tuned on the development set (and not even trained on the training set).
Consequently, there should be almost no difference between the scores on the test and development set.

Overall, the results show that both systems achieve between decent and good precision while clearly suffering from low recall.
Compared to Citron, the rule-based system has lower precision, but higher recall of \msg{} resulting in a slightly better F1 score.
For the roles, Citron has both better precision and better recall than RBS.
Together (joint measure of \msg{} and roles), Citron again surpasses RBS in both precision and recall.
As for predicting the type of a \msg{}, RBS has slightly higher recall, but greatly lower precision than Citron leading to slightly better F1 score of Citron.

For the rule-based system, the low recall mainly results from two causes.
First, the system is not capable of predicting certain types of speech (\rep{} and \frin{}) or roles (\adr{}) that are present in the dataset.
Second, the system was designed to prefer quality to quantity when automatically extracting quotations from large amounts of raw text.
As such, the system has a preference for precision over recall even for types of speech that it can predict.

For Citron, the low recall also has two reasons.
First, the system only predicts quotes that have a \cue{} -- but only 70\% of all \msgs{} have a \cue{}.
Second, the \cue{} recall itself is low because Citron's \cue{} classifier a) cannot detect  a multi-word \cue{} and b) was designed only for verbs as \cue{}.

While RBS should produce highly similar results on the test resp. dev set, there is a difference in the performance.
This deviation in precision, recall and F1 between the test and dev dataset for RBS can be solely attributed to natural variations in the data, e.g. the different quotations contained in the documents of the test resp. dev set.
The documents in the dev set contain quotations that happen to be more aligned with the rules implemented in RBS, thus reaching a slightly higher scores.

In summary, the data-driven machine learning approach clearly outperforms the rule-based system.
However, both existing systems do not provide a high recall level thereby motivating the need for our presented dataset to enable the creation of new systems providing higher recall (and more fine-grained annotations etc.).
From our experimentation with the systems, we believe that a machine learning approach actually designed for the annotation schema will significantly improve the recall to a usable level.

\subsection{Ablation Study}
To support our view and verify that the low recall of the Citron system is largely an effect of its \cue{} limitations, we performed additional experiments with modified versions of the dataset.
First, we remove any quotations that do not have a \cue{}, i.e. that cannot be predicted by Citron.
The results are shown in Table \ref{tab:results} (lower half) for the data \textit{dev cue} (445 quotations, 1340 roles) and \textit{test cue} (468 quotations, 1400 roles).
Second, we further removed any quotations that have a multi-word \cue{} since Citron internally is limited to a single word \cue{}.
We re-trained Citron with the new data. %
The results are in the same table with the data column \textit{dev 1 cue} (255 quotations, 768 roles) and \textit{test 1 cue} (300 quotations, 899 roles).

For Citron, the effect is as expected:
The recall for both quotes and roles significantly increases while precision slightly decreases, leading to increased F1 scores across the board.
Limiting the dataset to quotations with any \cue{}, Citron sees +7.5 recall, +7.5 F1 on the dev set resp. +7.6 recall, +7.5 F1 test set joint scores combining quotations and roles.
As a comparison, for our rule-based system precision decreases (-9.3/-6.9), recall increases (+2.6/+2.2) and F1 remains unchanged (+0.0/+0.2) for dev/test joint scores.
For limitation to a single-word \cue{}, RBS performance degrades as its rule set is no longer compatible with the artificially reduced dataset: Joint precision -21.1/-19.3, recall +3.2/+1.8 and F1 -6.6/-6.0.
Citron, however, improves another +8.7 recall, +5.5 F1 for the dev set resp. +7.7 recall, +4.7 F1 for the test set joint scores.
Together, this results in an increase of +16.2 recall, +13.0 F1 on the dev set resp. +15.3 recall, +12.2 F1 on the test set joint scores.
Thus, we confidently attribute a large part of Citron's low recall to its \cue{} limitations.

Most multi-word \cue{} expressions in the dataset are either past tense constructions or common idioms (see Section \ref{sec:cue}) that could be replaced by a single verb.
When manually examining the data, there is no inherent difference in the difficulty between quotations and roles used in single- or multi-word \cue{} expressions.
Consequently, we are confident a better suited system is capable to achieve strong results on our dataset and thereby create a system that can automatically extract quotations with attributions from German news articles.

\section{Use cases}
In this section, we outline envisioned use cases with project partners from the Digital Humanities and Computational Social Sciences.
We demonstrate how our dataset can help researchers work on their research questions.
Note that we do not intend the data (WIKINEWS articles) to be analyzed directly.
While this may be interesting for specific research targeting WIKINEWS, we intend our annotated resource to be used to train machine learning systems which, in turn, enable the automatic creation of annotations on other data sources, thereby making it useful for a wide range of applications.

Such a system can produce a list of quotations/speaker pairs from a collection of documents.
This allows researchers to quickly analyze the quotations contained in their data without laboriously reading every single document in a potentially large collection.
The system can further provide the \cue{} and \frm{} to automatically mark negations,  classify the type of quotation, aggregate quotations by their \cue{} word and provide statistics on these aspects.
Thereby, researchers can both have a quantitative view on the quotations in their data of interest as well as qualitatively analyze individual quotations and/or speakers by filtering all detected quotations for certain aspects.

For example, social climate science researchers can compare statements after grouping the speakers into politicians, environmental activists, corporate representatives and other public figures (this is possible after performing co-reference resolution and entity linking on the documents).

Another example is the comparison of different news outlets based on the general frequency resp. fraction of text being a quotation as well as the attributes of quotations used:
Type of quotation (e.g. \dir{} versus some \indi{} form), presence versus absence of a \src{}.
After collecting news articles on the same topic during the same timeframe for various media outlets, researchers can quantitatively compare the news outlets and analyze whether this correlates with news outlet metadata such as reach, geographical location, position in the political spectrum.
Further, it is possible to check if individual quotations occur in multiple news outlets or only once -- these cases could be candidates for a manual verification or otherwise of interest.

\section{Conclusion}
We presented a new dataset for quotation attribution in German news articles.
The dataset is freely available under a Creative Commons license and provides curated, high-quality annotations.
The fine-grained annotation schema allows the data to be used for various applications as it includes not only specifies who said what but also how, in which context, to whom and the type of quotation.

We described our annotation schema and dataset creation in detail, provided inter-annotator agreement and performed a quantitative analysis of the final dataset.
Finally, we evaluated two existing systems on our new dataset showing that a new approach is required to provide a high quality automatic detection of quotations.
While the systems managed to achieve an acceptable precision, they were only able to detect a subset of all annotations leading to low recall.

In the future, we want to create a system using the full potential of the dataset to be able to automatically obtain attributed quotations from news articles.

\section{Ethical Considerations and Limitations}
Automating tasks to scale to large data collections always carries a certain risk.
In the case of this paper, the dataset is the foundation to create a system that can extract attributed quotations from German news articles with high precision and recall (but certainly not error free).
Identifying who said what to whom according to news media on a large scale poses only a small risk compared to generating fake quotations (instead of extracting real ones) with already available state-of-the-art large language models.
Moreover, there are already existing rule-based systems (with precision and/or recall issues) to extract quotations and speakers automatically.
In any case, our dataset also provides the type of quotation so when using the identified quotations for a further analysis, it is possible to interpret the results more appropriately.

The dataset in itself is based on a freely available resource and uses a random sample without any focus on particular topics, speakers, authors or sources.
However, the articles in WIKINEWS might include certain biases as some articles will be written by the same authors, have the same source news agencies etc.
Being an open, collaborative platform, the raw articles should still be less biased than relying only on licensed articles from one specific news outlet.

Our annotations are likely neither perfectly error/bias free nor all-encompassing.
Sometimes, it is a balancing act to decide whether a certain sentence contains a quotation or the article author only phrased the sentence in a certain way to suggest a quotation.
Yet, we employed all means to create high-quality, fine-grained annotations to mitigate such issues by relying on skilled annotators, using annotation guidelines, weekly discussion meetings, curation and thorough handling of disagreements between annotators.

Overall, the possibility to extract who said what to whom according to news media can be an invaluable tool for researchers and journalists in their work to analyze the vast number of online media, help with identification of fake news based on their quotations or ease verification of quotations.

\section{Bibliographical References}\label{sec:reference}

\bibliographystyle{lrec-coling2024-natbib}
\bibliography{lrec-coling2024-statements,custom}

\end{document}